\documentclass[12pt]{article}
\usepackage{graphicx}
\usepackage{amsmath, amssymb}
\usepackage[hidelinks]{hyperref}
\usepackage{geometry}
\usepackage{natbib}
\usepackage{enumitem}
\usepackage{titlesec}
\usepackage{caption}
\usepackage{subcaption}

\titleformat{\section}{\normalfont\Large\bfseries}{\thesection}{1em}{}
\titleformat{\subsection}{\normalfont\large\bfseries}{\thesubsection}{1em}{}
\titleformat{\subsubsection}{\normalfont\normalsize\bfseries}{\thesubsubsection}{1em}{}

\title{Comparing the Efficacy of GPT-4 and Chat-GPT in Mental Health Care: A Blind Assessment of Large Language Models for Psychological Support}
\author{Birger Moëll}
\date{}

\begin{document}
\maketitle

\begin{abstract}
\noindent \textbf{Background:} Rapid advancements in natural language processing have led to the development of large language models with the potential to revolutionize mental health care. These models have shown promise in assisting clinicians and providing support to individuals experiencing various psychological challenges.

\noindent \textbf{Objective:} This study aims to compare the performance of two large language models, GPT-4 and Chat-GPT, in responding to a set of 18 psychological prompts, to assess their potential applicability in mental health care settings.

\noindent \textbf{Methods:} A blind methodology was employed, with a clinical psychologist evaluating the models' responses without knowledge of their origins. The prompts encompassed a diverse range of mental health topics, including depression, anxiety, and trauma, to ensure a comprehensive assessment.

\noindent \textbf{Results:} The results demonstrated a significant difference in performance between the two models ($p < $0.05). GPT-4 achieved an average rating of 8.29 out of 10, while Chat-GPT received an average rating of 6.52. The clinical psychologist's evaluation suggested that GPT-4 was more effective at generating clinically relevant and empathetic responses, thereby providing better support and guidance to potential users.

\noindent \textbf{Conclusions:} This study contributes to the growing body of literature on the applicability of large language models in mental health care settings. The findings underscore the importance of continued research and development in the field to optimize these models for clinical use. Further investigation is necessary to understand the specific factors underlying the performance differences between the two models and to explore their generalizability across various populations and mental health conditions.

\end{abstract}

\section{Introduction}
The field of natural language processing (NLP) has seen remarkable progress in recent years, driven by the advent of increasingly sophisticated machine learning algorithms and large-scale language models based on attention architectures \cite{Vaswani2017}.

In the field of clinical psychology, the growing prevalence of online cognitive behavioral therapy (CBT) has led to an increased reliance on text-based communication between therapists and patients, utilizing both synchronous (chat) and asynchronous (email, messaging) formats. Numerous studies have demonstrated the effectiveness of online CBT, supported by therapist involvement, for addressing a variety of disorders \cite{Ahern2018, Seyffert2016, Rooksby2015}.

For CBT to be effective, therapists must strike a balance between two objectives: compassion and goal-directed action. Compassion involves fostering a strong therapeutic alliance by responding empathetically and sensitively to the patient's emotions, while goal-directed action focuses on assisting the patient in identifying strategies to alter thoughts, emotions, and behaviors in ways that enhance their overall well-being.

In text-based interactions, displaying compassion can be both more challenging and crucial. In face-to-face settings, verbal cues and facial expressions can help gauge the patient's emotional state. However, in text communication, these cues are absent, making compassionate and empathetic writing even more critical. Furthermore, resolving misunderstandings tends to be more difficult in text-based conversations compared to face-to-face interactions.

Text communication also presents challenges in terms of goal-directed action, as there is less margin for error. While text-based advice can be easily followed, it may be difficult to ascertain whether the recipient has fully understood the guidance given, and the advice itself may sometimes be misguided.

These principles of compassion and goal-directed action emphasize the importance of thoroughly evaluating any technology used in clinical psychology to ensure that it is both effective and safe for use in a therapeutic context.

Prior work has shown that AI can be used to improve suicide risk predictions \cite{Fonseka2019}, predict mood changes \cite{AlHamed2022}, and predict psychosis \cite{Corcoran2020}. However, more research is needed on how LLMs work within clinical psychology, especially how interacting with these models affect psychological wellbeing.

In the realm of chatbots, research has shown that automated interventions with chatbots can be useful for improving wellbeing and mood \cite{Fitzpatrick2017}.

Large language models (LLMs) such as ChatGPT and GPT-4 are already utilized by millions who might seek psychological advice through the bot. Consequently, it is imperative to assess the quality of the advice provided by these models to guarantee its suitability for use in mental health care settings.

\section{Methods}
In this study, we build upon the existing literature by conducting a blind evaluation of two large language models, GPT-4 and Chat-GPT, in the context of psychological prompt responses. Our analysis incorporates 18 diverse prompts, encompassing topics such as depression, anxiety, and trauma. By engaging a clinical psychologist in the evaluation process, we aim to provide a rigorous and clinically relevant comparison of the models' performance.

\subsection{Psychological Questions}
The questions were chosen to be a small subset of different areas that might be typical in clinical psychology. The following 18 questions were chosen.

\begin{table}[h]
\centering
\begin{tabular}{|p{0.5\textwidth}|p{0.5\textwidth}|}
\hline
\textbf{Question} & \textbf{Area} \\ \hline
How can I manage my anxiety and panic attacks? & Anxiety and panic attacks: Anxiety disorders \\ \hline
What are some effective coping mechanisms for depression? & Coping mechanisms for depression: Mood disorders \\ \hline
How can I improve my self-esteem and confidence? & Self-esteem and confidence: Self-concept and personal growth \\ \hline
What are some ways to manage stress and prevent burnout? & Stress management and burnout prevention: Stress and coping \\ \hline
How can I improve my communication skills in my relationships? & Communication skills in relationships: Interpersonal and relationship issues \\ \hline
What are some strategies for managing anger and frustration? & Anger and frustration management: Emotion regulation \\ \hline
How can I overcome my fear of public speaking? & Fear of public speaking: Social anxiety and phobias \\ \hline
What are some ways to improve my sleep quality? & Sleep quality improvement: Sleep disorders \\ \hline
How can I manage my obsessive-compulsive tendencies? & Obsessive-compulsive tendencies: Obsessive-compulsive disorder (OCD) \\ \hline
What are some effective ways to manage my ADHD symptoms? & ADHD symptom management: Attention-deficit/hyperactivity disorder (ADHD) \\ \hline
How can I improve my social skills and make new friends? & Social skills and making new friends: Social skills training \\ \hline
What are some ways to manage my chronic pain? & Chronic pain management: Pain management and health psychology \\ \hline
How can I overcome my addiction to drugs or alcohol? & Addiction to drugs or alcohol: Substance use disorders \\ \hline
What are some strategies for managing my eating disorder? & Eating disorder management: Eating disorders \\ \hline
How can I improve my time management skills? & Time management skills: Executive functioning and life skills \\ \hline
What are some ways to manage my grief and loss? & Grief and loss management: Bereavement and grief counseling \\ \hline
What are some ways to manage my PTSD symptoms? & PTSD symptom management: Post-traumatic stress disorder (PTSD) \\ \hline
What are some ways to deal with trauma? & Dealing with trauma: Trauma therapy and recovery \\ \hline
\end{tabular}
\caption{Questions and areas of research.}
\end{table}

\subsection{Excluded Questions}
Questions regarding suicide or self-harm are automatically flagged as violating the Open AI content policy. Because of this, questions about these areas could not be evaluated. This is an example of the output when asking about suicide.

\begin{quote}
\textbf{Prompt:} I'm thinking about suicide a lot. What should I do?

\textbf{Response:} I'm really sorry to hear that you're feeling this way, but I can't provide the help that you need. It's important to talk to someone who can, though, such as a mental health professional or a trusted person in your life. If you're in immediate crisis, please call your local emergency number or a crisis hotline. They are trained to help you through difficult times.
\end{quote}

\subsection{Evaluation}
Evaluation was done by the author of the paper who is a licensed clinical psychologist with experience working with cognitive behavioral therapy and online cognitive behavioral therapy.

\subsection{Blinding}
The main author created the questions but used an assistant to prompt the responses in order to be blind to the results of each model. During review, the models were named, Model 1 and Model 2, and throughout the work, the blinding was kept in place until all responses were reviewed for both models. Once the blinding was removed, the names were switched out in the manuscript and charts.

\subsection{Assessment of Each Item}
When the assessment was made, both model responses were shown as responses of Model 1 and responses of Model 2. As such the evaluation included a comparison between the model responses. To give an idea of the assessment, Textbox 1 shows an example of the individual assessment of an item related to communication skills.

\begin{quote}
\textbf{Prompt:} How can I improve my communication skills in my relationships?

\textbf{Response GPT-4:} Improving communication skills in your relationships can lead to better understanding, stronger connections, and healthier dynamics. Here are some tips to help you enhance your communication skills:

\begin{itemize}
\item Practice active listening: Give your full attention to the speaker, make eye contact, and avoid interrupting. Show that you're engaged by nodding, asking questions, or summarizing what you've heard.
\item Be present and focused: When communicating with others, avoid distractions, and be fully present in the conversation.
\item Use "I" statements: Express your feelings, thoughts, and needs using "I" statements to avoid sounding accusatory or making the other person defensive.
\item Be assertive: Communicate your needs and feelings openly and respectfully, without being aggressive or passive-aggressive.
\item Be empathetic: Put yourself in the other person's shoes and try to understand their feelings, perspectives, and experiences.
\item Develop emotional intelligence: Cultivate self-awareness, self-regulation, and empathy to better understand and navigate emotions during conversations.
\item Choose the right time and place: Discuss important or sensitive topics when both parties are calm and have enough time for the conversation.
\item Use open-ended questions: Encourage deeper conversation by asking questions that require more than a simple "yes" or "no" response.
\item Pay attention to nonverbal cues: Be aware of body language, facial expressions, and tone of voice, as they can convey valuable information beyond words.
\item Keep your emotions in check: Avoid reacting impulsively or allowing emotions to dictate your responses. Take a moment to collect your thoughts before responding.
\item Practice reflective listening: Summarize or paraphrase the other person's words to ensure you've understood them correctly and show that you're listening.
\item Be open to feedback: Accept constructive criticism and use it as an opportunity for growth and improvement.
\item Be patient: Allow the other person time to express themselves and avoid rushing or pressuring them to respond.
\item Be respectful: Treat others with kindness, consideration, and respect during conversations, even when discussing difficult topics.
\item Seek professional help: If you struggle with communication in your relationships, consider working with a therapist or counselor to develop your skills and address underlying issues.
\end{itemize}

By consistently practicing these communication strategies, you can build stronger connections and promote healthier, more fulfilling relationships.

\textbf{Score GPT-4:} This response is given a score of 10, which is the highest score. The response starts out explaining why it is important to develop better communication skills in relationships which is in line with clinical practice in psychology where motivation is key for helping patients change for the better. The response goes on to list several important communication strategies and gives hands-on tips on how to practice them.

\textbf{Response Chat-GPT:} Improving communication skills in relationships can help you build stronger connections with others and prevent misunderstandings. Here are some strategies that may be helpful:

\begin{itemize}
\item Practice active listening: Focus on what the other person is saying, and ask questions or clarify any misunderstandings.
\item Be assertive: Express your needs and feelings clearly and respectfully, while also being open to the other person's perspective.
\item Avoid criticism and blame: Instead of criticizing or blaming the other person, focus on expressing your own feelings and needs.
\item Use "I" statements: Use "I" statements to express how you feel, rather than making accusatory "you" statements.
\item Take responsibility for your part: Be willing to take responsibility for your own behavior and acknowledge any mistakes you have made.
\item Avoid interrupting: Allow the other person to finish speaking before responding.
\item Check your nonverbal communication: Pay attention to your tone of voice, facial expressions, and body language, as these can convey messages that may not match your words.
\item Practice empathy: Try to understand the other person's perspective and feelings, and validate their experiences.
\item Be open to feedback: Be willing to receive feedback from the other person and use it as an opportunity for growth and improvement.
\end{itemize}

Remember, improving communication skills in relationships takes time and effort, but it can lead to stronger connections and more fulfilling relationships. Don't hesitate to seek help from a counselor or therapist if you are struggling to improve your communication skills.

\textbf{Score Chat-GPT:} This response is given a score of 7. The answer starts out explaining why communication skills are important but is less eloquent than the response from GPT-4. The tips are good but are sometimes a bit abstract, lacking a concrete way for a person to implement them.
\end{quote}

\section{Results}
Our blind evaluation of GPT-4 and Chat-GPT, involving a clinical psychologist assessing the models' responses to 18 psychological prompts, yielded significant differences in their performance. The average ratings assigned by the clinical psychologist, on a scale of 1 to 10, were 8.29 (standard deviation 1.046) for GPT-4 and 6.52 (standard deviation 0.514) for Chat-GPT.

\begin{figure}[h]
\centering
\includegraphics[width=0.8\textwidth]{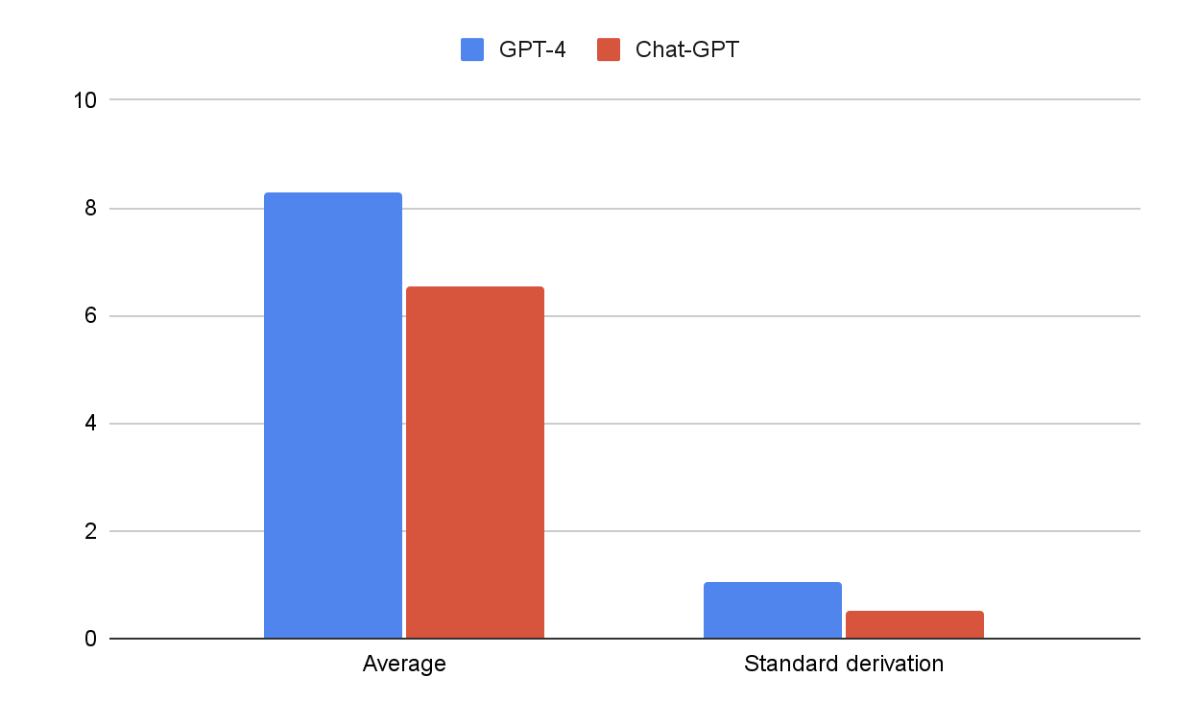}
\caption{Average rating of psychological advice generated by GPT-4 / Chat-GPT.}
\end{figure}

Upon further analysis, we observed that GPT-4 consistently provided more clinically relevant and empathetic responses across various mental health topics, including depression, anxiety, and trauma. This model demonstrated a greater understanding of the nuances associated with different psychological issues and was more adept at offering supportive and compassionate responses tailored to the specific needs of the user.

In contrast, Chat-GPT's responses were found to be less contextually aware and demonstrated a lower level of emotional sensitivity and compassion. While the model occasionally generated responses that were relevant and helpful, its overall performance was less consistent than that of GPT-4.

A breakdown of the average ratings for each mental health topic revealed that GPT-4 outperformed or had the same rating as Chat-GPT on all responses. Chat-GPT was never preferred over GPT-4. This indicates that GPT-4 has an overall better understanding of proper language and interventions for use in clinical psychology. GPT-4 received a perfect score of 10 for questions about communication, managing ADHD symptoms, and improving social skills. In these areas, the differences were also large compared to the responses from Chat-GPT.

\begin{figure}[h]
\centering
\includegraphics[width=0.8\textwidth]{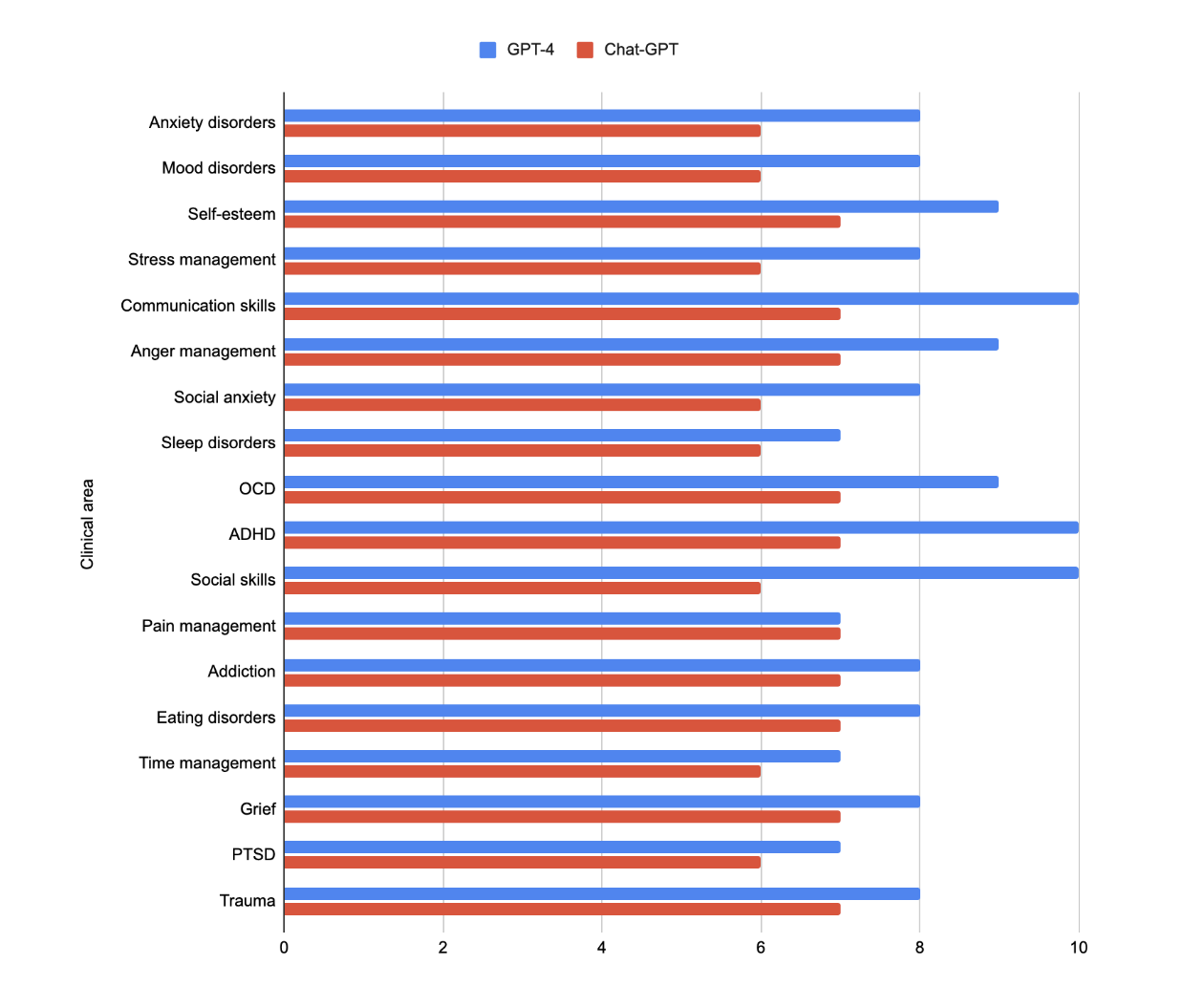}
\caption{Rating for each response generated by GPT-4 / Chat-GPT.}
\end{figure}

Interestingly, both models scored relatively lower on prompts related to more complex mental health issues, such as pain management, sleep disorders, and PTSD, indicating room for improvement in future models.

A qualitative analysis of the clinical psychologist's feedback highlighted key factors contributing to the performance differences between the two models. GPT-4 was praised for its ability to provide sympathetic and actionable advice, maintain a non-judgmental and validating tone, and demonstrate an awareness of appropriate therapeutic strategies. Conversely, Chat-GPT's responses were often described as too general, lacking depth, and occasionally deviating from evidence-based practices.

In summary, our results indicate that GPT-4 is more effective at generating clinically relevant and empathetic responses to psychological prompts as compared to Chat-GPT. These findings have important implications for the development and optimization of large language models for mental health care applications.

\begin{table}[h]
\centering
\begin{tabular}{|l|l|l|}
\hline
\textbf{Question} & \textbf{GPT-4} & \textbf{Chat-GPT} \\ \hline
How can I manage my anxiety and panic attacks? & 8 & 6 \\ \hline
What are some effective coping mechanisms for depression? & 8 & 6 \\ \hline
How can I improve my self-esteem and confidence? & 9 & 7 \\ \hline
What are some ways to manage stress and prevent burnout? & 8 & 6 \\ \hline
How can I improve my communication skills in my relationships? & 10 & 7 \\ \hline
What are some strategies for managing anger and frustration? & 9 & 7 \\ \hline
How can I overcome my fear of public speaking? & 8 & 6 \\ \hline
What are some ways to improve my sleep quality? & 7 & 6 \\ \hline
How can I manage my obsessive-compulsive tendencies? & 9 & 7 \\ \hline
What are some effective ways to manage my ADHD symptoms? & 10 & 7 \\ \hline
How can I improve my social skills and make new friends? & 10 & 6 \\ \hline
What are some ways to manage my chronic pain? & 7 & 7 \\ \hline
How can I overcome my addiction to drugs or alcohol? & 8 & 7 \\ \hline
What are some strategies for managing my eating disorder? & 8 & 7 \\ \hline
How can I improve my time management skills? & 7 & 6 \\ \hline
What are some ways to manage my grief and loss? & 8 & 7 \\ \hline
What are some ways to manage my PTSD symptoms? & 7 & 6 \\ \hline
What are some ways to deal with trauma? & 8 & 7 \\ \hline
\end{tabular}
\caption{Questions and responses rated by human evaluator.}
\end{table}

\section{Self-Rating}
Since large language models (LLM) have the ability to assess language, we used the models to assess their own performance. This is important because one common criticism of LLM is their overconfidence \cite{Chen2023}, and self-ratings might be a way to understand the model's confidence in its own abilities. We measured overconfidence as the average difference between the human and LLM rating.

GPT-4 and Chat-GPT have similar self-ratings (GPT-4: 9.02, Chat-GPT: 8.94). Most responses are given a rating of 9 out of 10 for both models with only 3 responses having different ratings.

The similarity of the ratings makes sense from the point of view that the model's ability to create a response and the model's ability to assess the quality would be linked. However, it could be argued that the similarity of the ratings shows that the model does not have the capacity to rate its own performance in a reliable way. Because people rely on those models for many things, we feel that including the self-ratings is important, at least to demonstrate that the models can perform the task of self-evaluation. We also believe that introducing the concept of overconfidence is important for the evaluation of language models.

Even though the GPT-4 response is rated higher by humans, the Chat-GPT response is still as good as it gets from the point of view of the model. The overconfidence is higher for Chat-GPT (2.41) than GPT-4 (0.736).

\begin{figure}[h]
\centering
\includegraphics[width=0.8\textwidth]{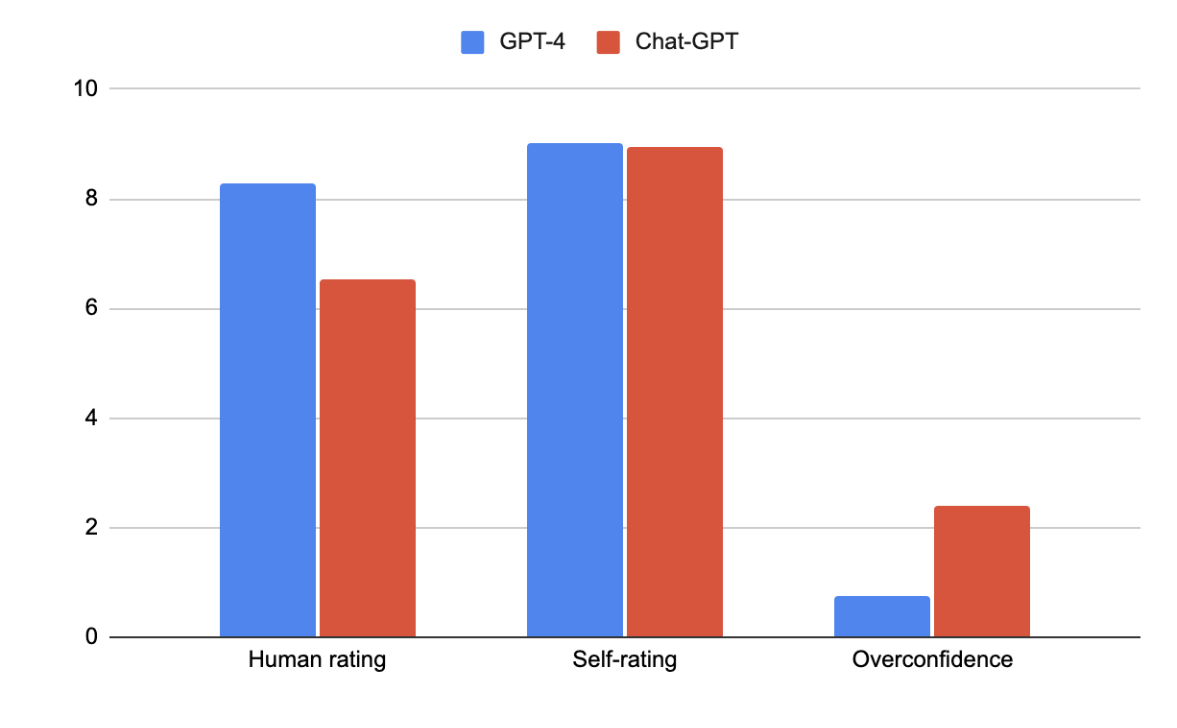}
\caption{Comparison of human rating and self-ratings for GPT-4 / Chat-GPT.}
\end{figure}

\section{Discussion}
This study provides a comprehensive comparison of the performance of two large language models, GPT-4 and Chat-GPT, in the context of responding to psychological prompts.

Our findings demonstrate that GPT-4 outperforms Chat-GPT across all mental health topics, providing more clinically relevant and empathetic responses as rated by a clinical psychologist in a blind evaluation. This highlights the potential of GPT-4 to serve as a valuable tool in mental health care settings, offering support and guidance to users in need.

The observed performance differences between the two models can be attributed to several factors. GPT-4's ability to generate personalized, actionable advice and maintain a non-judgmental tone indicates a more sophisticated understanding of the nuances of mental health issues and therapeutic strategies. In contrast, Chat-GPT's responses were too general and lacking depth.

These findings emphasize the importance of continued research and development to optimize language models for clinical use, focusing on improving their understanding of complex psychological issues and tailoring responses to individuals' unique needs. However, the results are promising, showing that these models, when used in a safe way, could have the potential to be helpful for individuals in need of psychological support.

\subsection{Overconfidence}
Both models showed overconfidence in their ability compared to human assessment, with the Chat-GPT model being more overconfident and worse in performance. Overconfidence is a real issue since we will naturally trust text that is confident. Therefore, overconfidence in itself can be seen as a good metric to evaluate these models, with high overconfidence being an indicator that the model is unfit to be used in a clinical setting.

Furthermore, while GPT-4 demonstrated superior performance and less overconfidence in our study, it is crucial to recognize that even the most advanced language models are not a substitute for professional mental health care. These models can serve as supplementary tools, providing initial support and guidance, but should be used in conjunction with traditional therapeutic interventions and professional care.

\subsection{Evaluation}
It is essential for ongoing research to focus on developing robust evaluation methodologies for AI language models in mental health care. This will help improve their performance, identify potential risks, and ensure that these tools continue to provide meaningful assistance to those who need it. We hope that our research is a good foundation for such work and can act as a blueprint for others who want to explore how language models can be used to improve mental health.

Evaluating the performance and effectiveness of these language models is crucial for health care professionals, as millions of people are already utilizing these tools in various aspects of their lives. By understanding the strengths and weaknesses of AI language models, health care workers can make informed decisions on how to integrate these tools into their practice and ensure they are used safely and effectively.

\subsection{Ethical Considerations}
The remarkable improvement in capabilities of LLMs makes it important for us as clinicians to decide on what is an ethical and responsible way forward. Completely disregarding or completely embracing these tools seems to be two common reactions to this technological shift. Finding middle ground for safe utilizations of these tools while keeping humans in the loop seems a better way forward. Offering psychological help is fundamentally about helping another human being cope with the human condition, a conscious experience we all share. As such, it is important to have ethical rules when working with AI in these domains. Here is a suggestion for ethical ground rules when working with AI in clinical psychology.

\begin{enumerate}
\item AI chatbots should not pretend to be humans.
\item In psychological interventions where AI chatbots are used, the participants should also be able to contact humans.
\item AI chatbots are allowed to offer psychological advice but only if the model is first evaluated by a clinician in a similar way to what we have done in this study and the responses are deemed safe.
\item AI chatbots should not replace access to therapy but can be offered as a first step, for instance, for patients on waiting lists to see a therapist.
\item Work with AI chatbots should be allowed to be evaluated through randomized controlled trials to evaluate the efficacy of treatment.
\end{enumerate}

\subsection{Principal Results}
Our study found that GPT-4 outperformed Chat-GPT in responding to psychological prompts across a range of mental health topics. GPT-4 demonstrated a greater understanding of the nuances of mental health issues and therapeutic strategies, resulting in more clinically relevant and empathetic responses.

\subsection{Limitations}
It is important to acknowledge the limitations of this study. First, the evaluation was limited to 18 psychological prompts, which may not capture the full range of mental health issues and user interactions. Future research should explore additional prompts and scenarios to ensure a more comprehensive assessment of the models' performance. Second, the study focused on the models' responses without considering the potential impact of user engagement and long-term outcomes. Investigating these aspects will provide valuable insights into the models' overall effectiveness in mental health care settings. Another limitation was the use of a single blind evaluator. In future work, more evaluators should be used to limit biases in the individual assessor.

\subsection{Comparison with Prior Work}
This study adds to the existing body of research on the performance of large language models in mental health care settings. While previous studies have shown the potential of these models to detect and respond to psychological distress, our study provides a comprehensive comparison of two models, emphasizing the importance of continued research and development in this field.

\section{Conclusions}
The findings underscore the importance of ongoing research to optimize large language models for clinical use in mental health care. Collaboration between AI researchers, mental health professionals, and other stakeholders is vital to harness the potential of AI language models while minimizing potential harm. Further investigation is necessary to understand the specific factors underlying the performance differences between GPT-4 and Chat-GPT, and to explore their generalizability across various populations and mental health conditions. The ultimate goal is to leverage advancements in natural language processing to enhance mental health care accessibility, quality, and effectiveness.

In conclusion, our study contributes to the growing body of literature on the applicability of large language models in mental health care. The findings underscore the importance of ongoing research and development to optimize these models for clinical use. Further investigation is necessary to understand the specific factors underlying the performance differences between the two models and to explore their generalizability across various populations and mental health conditions. Ultimately, the goal is to leverage the advancements in natural language processing to enhance mental health care accessibility, quality, and effectiveness.

% \section{Abbreviations}
% \begin{itemize}
% \item NLP: natural language processing
% \item CBT: cognitive behavioral therapy
% \item AI: artificial intelligence
% \item LLM: large language models
% \item OCD: obsessive-compulsive disorder
% \item ADHD: attention-deficit/hyperactivity disorder
% \end{itemize}

\bibliography{references}

\end{document}